\documentclass[journal]{IEEEtran}
\usepackage{cite}

\ifCLASSINFOpdf
  \usepackage[pdftex]{graphicx}
  \graphicspath{{../pdf/}{../jpeg/}}
  \DeclareGraphicsExtensions{.pdf,.jpeg,.png}
\else
\fi
\usepackage{amsmath}
\usepackage{array}

 \usepackage[caption=false,font=footnotesize]{subfig}
\usepackage{url}

\usepackage{amsmath}
\usepackage{amsthm}
\usepackage{booktabs}
\usepackage{multirow}
\usepackage{amssymb}
\usepackage{algorithm}
\usepackage{algorithmic}

\begin{document}
%
\title{ Complementary Pseudo Labels For Unsupervised Domain Adaptation On Person Re-identification }
%
%
%

\author{Hao~Feng,
        Minghao~Chen,
        Jinming~Hu,
        Dong~Shen,
        Haifeng~Liu,
        Deng~Cai,~\IEEEmembership{Member,~IEEE}
\thanks{This work was supported in part by The National Key Research and Development Program of China (Grant Nos: 2018AAA0101400), in part by The National Nature Science Foundation of China (Grant Nos: 62036009, U1909203, 61936006, 61973271), in part by the Alibaba-Zhejiang University Joint Institute of Frontier Technologies.\textit{(Corresponding author: Deng~Cai.)}}%
\thanks{Hao~Feng, Minghao~Chen, Jinming~Hu, Dong~Shen, Deng~Cai are with the State Key Laboratory of Computer-Aided Design (CAD) and Computer Graphics (CG), Zhejiang University, HangZhou 310027, China.(email: fenghao9523@gmail.com)}%
\thanks{Deng Cai is also with the Alibaba-Zhejiang University Joint Institute of Frontier Technologies, 866 Yuhangtang Road, Xihu District, Hangzhou 310027, China.(email: dengcai@gmail.com)}%
\thanks{Haifeng Liu is with the College of Computer Science, Zhejiang University, Hangzhou, China, 310027.}%
}

%
%

\markboth{Journal of \LaTeX\ Class Files}%
{Shell \MakeLowercase{\textit{et al.}}: Bare Demo of IEEEtran.cls for IEEE Journals}
%



\maketitle

\begin{abstract}
  In recent years, supervised person re-identification (re-ID) models have received increasing studies. However, these models trained on the source domain always suffer dramatic performance drop when tested on an unseen domain. Existing methods are primary to use pseudo labels to alleviate this problem. One of the most successful approaches predicts neighbors of each unlabeled image and then uses them to train the model. Although the predicted neighbors are credible, they always miss some hard positive samples, which may hinder the model from discovering important discriminative information of the unlabeled domain. In this paper, to complement these low recall neighbor pseudo labels, we propose a joint learning framework to learn better feature embeddings via high precision neighbor pseudo labels and high recall group pseudo labels. The group pseudo labels are generated by transitively merging neighbors of different samples into a group to achieve higher recall. However, the merging operation may cause subgroups in the group due to imperfect neighbor predictions. To utilize these group pseudo labels properly, we propose using a similarity-aggregating loss to mitigate the influence of these subgroups by pulling the input sample towards the most similar embeddings. Extensive experiments on three large-scale datasets demonstrate that our method can achieve state-of-the-art performance under the unsupervised domain adaptation re-ID setting.
\end{abstract}


%
\IEEEpeerreviewmaketitle

\section{Introduction}
%
%
%
%

\IEEEPARstart{P}{erson} re-identification (re-ID) \cite{DBLP:journals/corr/ZhengYH16} aims to find the same pedestrian across different cameras with a given query image. Although recent supervised re-ID methods have made significant progress \cite{DBLP:conf/iccv/ChenDXYCYRW19,DBLP:conf/cvpr/ZhengYY00K19,DBLP:journals/tip/WeiWJYHCHH20,DBLP:conf/cvpr/0004GLL019,DBLP:journals/corr/abs-2006-02631,DBLP:conf/eccv/HeL20}, the models trained on the source domain have a considerable performance drop when transferred to an unseen domain. To mitigate this problem, unsupervised domain adaptation (UDA) is one of the most feasible solutions because it is easy to obtain enough unlabeled images in the target domain but expensive to annotate all these images.

Traditional UDA methods mainly focus on generic classification tasks, such as image classification and semantic segmentation. These tasks implicitly assume the source domain and the target domain share the same label space. Therefore, they prefer to align the feature distribution of different domains by utilizing the classifier \cite{DBLP:conf/icml/LongC0J15,DBLP:journals/corr/HoffmanWYD16,DBLP:conf/cvpr/YanDLWXZ17,DBLP:conf/iccv/0001XC19}. However, re-ID is an open-set problem, which means different domains have entirely different identities. Thus, the traditional aligning feature method is not suitable to re-ID. 

\begin{figure}[t]
	\centering
	\includegraphics[height=0.27\textwidth,width=0.41\textwidth]{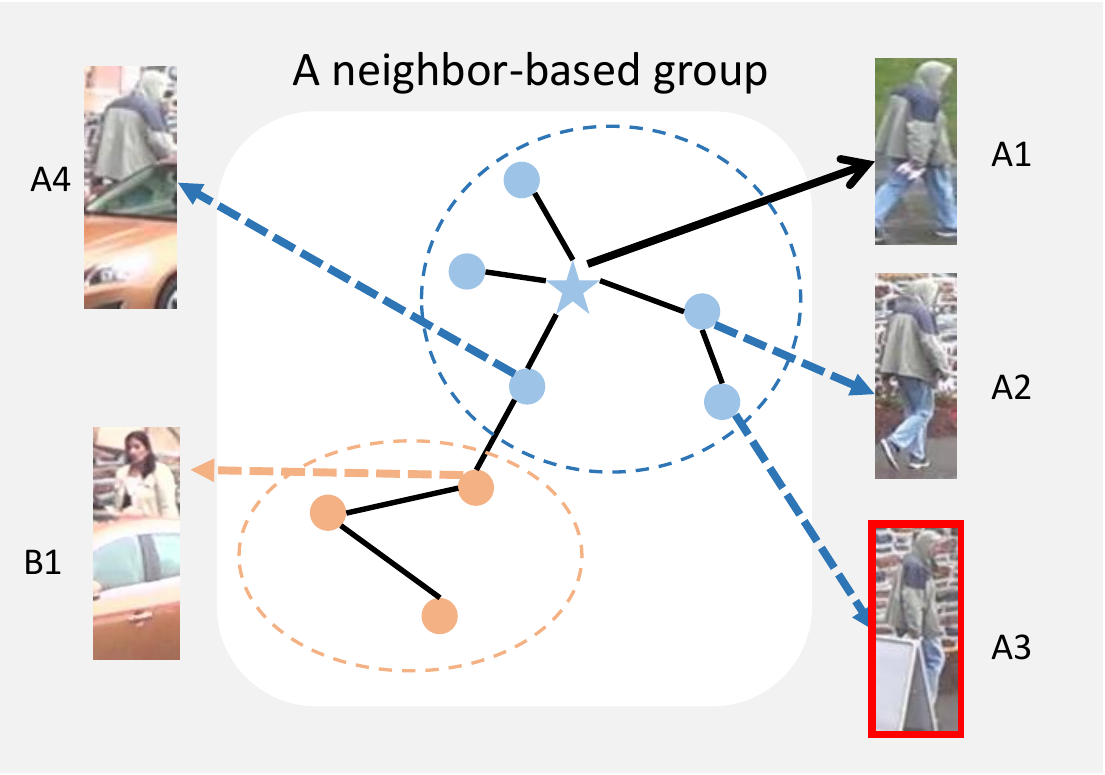}
	\caption{Illustration of neighbors and a merged group. Circles indicate different images. Circle colors denote person identities. Given the input image A1 (denoted by $\star$), its neighbors are those images connected by lines. Image A3 has the same identity as A1, but it is difficult to recognize it as a neighbor of A1 due to the background and occlusion. However, we can find A3 by A2. Both A4 and B1 are severely occluded by a car, so they are easily predicted to share the same identity. Merging B1 into the group of identity A naturally causes subgroups. }
	\label{fig:intro}
\end{figure}
 
Recently, some studies on the UDA problem in re-ID have achieved some promising improvements by using pseudo labels to learn the discriminative information of the target domain \cite{DBLP:journals/pr/SongWZDZHW20,DBLP:conf/iccv/FuWWZSUH19,DBLP:conf/cvpr/Zhong0LL019}. They utilize the discriminability of the source domain to assign pseudo labels in the target domain and then train the model with these pseudo labels. Among these methods, some cluster-based methods \cite{DBLP:journals/pr/SongWZDZHW20,DBLP:conf/iccv/FuWWZSUH19} use DBSCAN \cite{DBLP:conf/kdd/EsterKSX96} to divide target samples into different clusters and generate pseudo labels. Despite state-of-the-art performance reported by them, these methods are still restricted by the inevitable random label noise. Besides, the property of the label noise is hard to know due to the clustering algorithm. In addition to these cluster-based methods, there exists a state-of-the-art neighbor-based method \cite{DBLP:journals/corr/abs-1908-00485} that assigns pseudo labels based on neighbors which are predicted by graph convolution network(GCN) \cite{DBLP:conf/cvpr/Wang0LW19}. These neighbors are viewed as multi-class pseudo labels to train the model. Although the predicted neighbors are credible, these neighbors always ignore some hard positive samples. Specifically, given a target sample, the samples that share the same identity but with different camera views or occlusions are hard to be included in the neighbors, as shown in Figure. \ref{fig:intro}. These missing hard samples will undoubtedly hinder the model from robust with different views and occlusions but are mostly ignored by the neighbor-based UDA method.

To mitigate the problem of missing hard samples in the neighbor-based UDA method, we propose a joint learning framework. Our framework can discover and exploit more hard positive samples by jointly optimizing the model using complementary pseudo labels, \textit{i.e.} high precision neighbor pseudo labels and relative higher recall group pseudo labels. In this paper, we first make a basic assumption that two different unlabeled target domain person images belong to the same identity if they share common neighbors. We start by collecting target domain images' neighbors at each training iteration. According to the above assumption, we transitively merge all target domain images' neighbors to divide all the images into different groups. Images in the same group are assigned the same pseudo label. Obviously, the merging operation lets a group contain more samples with the same identity than neighbors, which means a higher recall. The learned feature embeddings can be more robust and discriminative by training the model with these group pseudo labels.

However, it is not naive to utilize the group pseudo labels properly. The predicted neighbors are not perfect and may contain some negative samples with a different identity due to similar backgrounds or frequent pedestrian occlusions. Merging the neighbors of these negative samples into one group makes the group noisy. Thus a group may contain multiple subgroups corresponding to multiple identities, as shown in Figure. \ref{fig:intro}. This inherent structure of the merged group is the main difference with clusters generated by DBSCAN \cite{DBLP:conf/kdd/EsterKSX96}. Considering subgroups in the merged group, we hope the input sample to be closer to the most similar subgroup than other subgroups to mitigate their influence. Hence we introduce a similarity-aggregating loss based on the assumption that with a good embedding function, the embeddings which share the same identity should be closer than the embeddings with different identities. The introduction of similarity-aggregating loss to utilize the group pseudo labels is also the key to our framework's performance improvement.

In summary, our main contributions are as follows:
\begin{itemize}
\item We propose to tackle the missing hard samples problem in neighbor-based UDA methods for person re-ID by using a joint learning framework with high precision neighbor pseudo labels and relative higher recall group pseudo labels. Experiments show the mutual benefit between these complementary pseudo labels.
\item Conventional softmax loss cannot utilize group pseudo labels properly due to the existence of subgroups. Hence we introduce a similarity-aggregating loss to utilize group pseudo labels properly. Experiments demonstrate that the similarity-aggregating loss is more effective with group pseudo labels.
\item Our method achieves state-of-the-art unsupervised domain adaptation performance on three popular person re-ID benchmarks: Market-1501, DukeMTMC-reID and MSMT17.
\end{itemize}

\begin{figure*}[t]
  \centering
  \includegraphics[height=0.48\textwidth,width=0.95\textwidth]{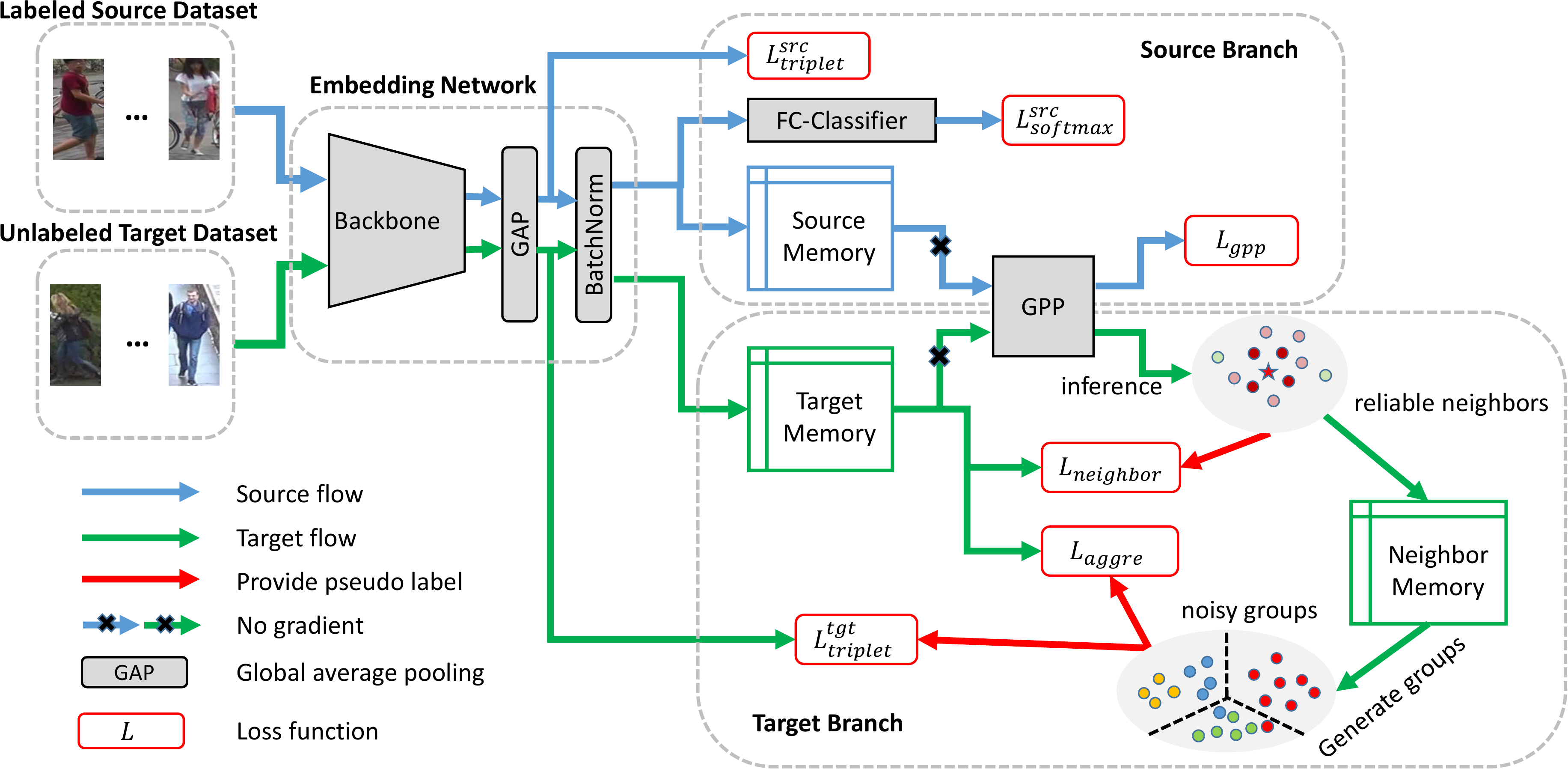}
  \caption{Overview of our joint learning framework. We sample inputs from the source dataset and the target dataset and obtain their embeddings by the embedding network. The source embeddings are utilized in a supervised manner with softmax cross-entropy loss $L_{softmax}^{src}$ and triplet loss $L_{triplet}^{src}$. The target embeddings are used with the target memory module that stores target embeddings. The graph-based positive prediction (GPP) network trained on the source dataset ($L_{gpp}$) infers the reliable neighbors (highlighted) of the input target sample (denoted by $\star$). Then we generate noisy groups by transitively merging all the reliable neighbors. Using reliable neighbors, we compute the neighbor-recognizing loss $L_{neighbor}$. We calculate similarity-aggregating loss $L_{aggre}$ and target triplet loss $L_{triplet}^{tgt}$ with the group pseudo labels.}
  \label{fig:framework}
\end{figure*}

\section{Related work}
\subsection{Unsupervised Domain Adaptation} 
Unsupervised domain adaptation aims to transfer the knowledge of the source domain to the target domain. The core idea of conventional UDA is to learn domain invariant features. To achieve this goal, some Maximum Mean Discrepancy (MMD) based methods \cite{DBLP:conf/icml/LongC0J15,DBLP:journals/corr/TzengHZSD14,DBLP:conf/cvpr/SaitoWUH18} and adversarial learning based methods \cite{DBLP:conf/cvpr/ZhangO0018,DBLP:conf/cvpr/TzengHSD17,DBLP:journals/jmlr/GaninUAGLLML16} have been proposed to minimize the feature distribution divergence. There are also some methods based on pseudo-labels. \cite{DBLP:conf/icann/DasL18,DBLP:conf/icml/LiangHF20} both apply pseudo-labeling for the unlabelled target domain data to refine the model. However, these methods assume that the source domain and the target domain share the same class labels, while different domains have entirely different identities in re-ID.

\subsection{Unsupervised re-ID}
UDA is closely related to unsupervised learning in re-ID. \cite{DBLP:conf/aaai/LinD00019} introduces a bottom-up clustering framework to train the network by clustering and merging. \cite{DBLP:journals/tip/YeLMZY19} proposes a Dynamic Graph Matching (DGM) framework to refine the label estimation process by learning a better distance measurement. \cite{9154587} utilizes the instance-wise supervision to train a data augmentation invariant and instance spreading feature. However, these methods cannot achieve comparable performance as UDA methods due to the lack of source domain label knowledge.

\subsection{Unsupervised Domain Adaptation re-ID}
Recent UDA re-ID studies mainly focus on two categories: reducing the domain gap and directly learning the unlabeled domain's discriminative information. The first category mostly tries to translate the source domain images to the target domain to reduce the domain gap in a common space. PTGAN \cite{DBLP:conf/cvpr/WeiZ0018} considers the person foregrounds constraints to ensure the stability of their identities after translating. Similarly, SPGAN \cite{DBLP:conf/cvpr/Deng0YK0J18} proposes the self-similarity constraint and the domain-dissimilarity constraint to generate samples that possess the target domain style while preserving their ID information. Furthermore, ATNet \cite{DBLP:conf/cvpr/LiuZCH019} decomposes the domain gap into a set of essential factors to transfer styles more precisely. Despite their effectiveness, these methods still leave a significant performance gap compared with supervised approaches. 

The second category uses pseudo labels to learn discriminative information of the unlabeled domain. \cite{DBLP:journals/pr/SongWZDZHW20} utilizes DBSCAN \cite{DBLP:conf/kdd/EsterKSX96} to discover clusters of the target domain without knowing the number of identities. These clusters can be viewed as pseudo labels to facilitate training. SSG \cite{DBLP:conf/iccv/FuWWZSUH19} extends the previous method by exploiting the potential similarity of the global body and local parts to build multiple clusters. \cite{DBLP:journals/corr/abs-1908-00485} introduces a GCN to explore each target sample's reliable neighbors as pseudo labels. \cite{DBLP:conf/cvpr/YuZWGGL19} compares the target sample with a set of known source domain reference samples to generate the soft multilabel. Although these methods achieve high performance using different pseudo labels, they neglect noise pseudo labels. Therefore these methods cannot achieve optimal performance.

\subsection{Learning with Noisy Labels}
We use the similarity-aggregating loss to mitigate the label noise of the group pseudo labels. There are also some studies about learning with noisy labels. \cite{DBLP:conf/cvpr/PatriniRMNQ17} estimate the noise transition matrix to correct labels. Generalized Cross Entropy loss \cite{DBLP:conf/nips/ZhangS18} combines mean-absolute loss and cross-entropy loss to be robust against noisy labels. There are also some methods \cite{DBLP:conf/iclr/GeCL20,DBLP:conf/aaai/YangLZLSCGHJL20} designed for unsupervised cross-domain person re-ID. \cite{DBLP:conf/aaai/YangLZLSCGHJL20} designs an asymmetric co-teaching framework for pseudo labels generated by DBSCAN \cite{DBLP:conf/kdd/EsterKSX96}. \cite{DBLP:conf/iclr/GeCL20} mitigates the pseudo label noise by using soft triplet labels. However, these above methods are all designed for random noise, while the noise of our group pseudo labels is mainly caused by the subgroups.

\section{Proposed Methods}
\subsection{Problem Definition}
UDA in re-ID assumes that we have a labeled source domain with a set of $N_s$ images $X_s=\{x_1^s,x_2^s,...,x_{N_s}^s\}$ and an unlabeled target domain with a set of $N_t$ images $X_t=\{x_1^t,x_2^t,...,x_{N_t}^t\}$. Each image $x_i^s$ in the source domain has a corresponding identity $y_i^s$, where $y_i^s \in \{1,2,...,M\}$. $M$ is the number of identities in the source domain. Note that the identity annotation of each image $x_i^t$ in the target domain is not available. Our goal is to learn a robust feature embedding network $f(\theta)$ on the unlabeled target domain. We denote the embeddings of the source images and target images as $F^s=\{f_1^s,f_2^s,...,f_{N_s}^s\}$ and $F^t=\{f_1^t,f_2^t,...,f_{N_t}^t\}$ respectively. In the testing stage, we use the embedding of the query image to find the same pedestrian in the gallery images.

To tackle the UDA problem, we propose a joint learning framework using complementary pseudo labels. Our key idea is to optimize the feature embedding network with high precision neighbor pseudo labels and high recall group pseudo labels simultaneously. Besides, the group pseudo labels may have subgroups and are hard to utilize. Therefore, we use a similarity-aggregation loss to achieve optimal unsupervised domain adaptation performance.

\subsection{Neighbor-based UDA Method Revisit} \label{section:revisit}
Neighbor-based UDA methods predict positive neighbors of each target image, which share the same identity, to supervise the model training as the pseudo labels. The most successful method \cite{DBLP:journals/corr/abs-1908-00485} of them proposes a graph-based positive prediction (GPP) network to improve the neighbor prediction. The GPP network consists of several graph convolution layers and a binary classifier and takes the $k$-nearest-neighbors $K(x_i^t)$ of $x_i^t$ as inputs. The $k$-nearest-neighbors are selected according to the cosine similarities between $f_i^t$ and $F^t$. The GPP network predicts the probability that each nearest-neighbor has the same identity with the given sample $x_i^t$. Finally, the predicted neighbors of $x_i^t$ are defined as,
\begin{equation}
  \Omega_i=\{j|j \in K(x_i) \wedge p_j^* \geq \mu \},
\end{equation}
where $\mu$ is a threshold value, and $p^*$ is the probability predicted by the GPP network. These neighbors $\Omega_i$ of $x_i^t$ can be viewed as a multi-class pseudo label to recognize $x_i^t$ to $j \in \Omega_i$, named neighbor pseudo labels. To avoid computing $F^t$ at each training batch, they maintain a memory bank which stores a running average $v_i$ for each $f_i^t$, denoted as $V=\{v_1,v_2,...,v_{N_t}\}$. The GPP network is optimized on the labeled source domain $\{x_i^s, y_i^s\}$ by binary cross-entropy loss $L_{gpp}$. Meanwhile, the feature embedding network $f(\theta)$ is optimized by neighbor pseudo labels $\Omega_i$ in the form of,
\begin{align}
L_{neighbor} &=  - \sum\limits_{j} w_{ij} \log p_{ij}, j \in \Omega_i, \notag \\
where\ w_{ij} &=
\begin{cases}
  \frac{1}{|\Omega_i|}, &  j \neq i  \\ 
  1, & j = i 
\end{cases}, \forall j \in \Omega_i, \\
and\ p_{ij} &= p(j|x^t_i) = \frac{\exp ({v_j}^\mathrm{T} f_i / \tau)}{\sum_{k=1}^{N_t} \exp ({v_k}^\mathrm{T} f_i / \tau))}, \notag \\
\end{align}
and $|\Omega_i|$ means the size of $\Omega_i$, $f_i$ denotes $f_i^t$ for simplicity. Here we name $L_{neighbor}$ as \textbf{neighbor-recognizing loss} because this loss recognizes the target image $x^t_i$ as its reliable neighbors, and the $w_{ij}$ can be viewed as a soft label.
 
However, the neighbor pseudo labels generated by GPP have low recall \cite{DBLP:journals/corr/abs-1908-00485} due to the imperfection of features and the manual threshold, which hinder the model robust from some hard scenes, such as view changing and occlusions. To make up for the low recall neighbor pseudo labels, we propose the joint learning framework and the similarity-aggregation loss to explore and utilize more hard samples.

\subsection{ Joint Learning Framework }
According to the experiments of section IV-E-3, the neighbors $\Omega_i$ predicted by the GPP network can achieve high precision but low recall. It means that the GPP network can discover highly similar samples correctly. However, it often ignores some hard positive samples that share the same identity but hard to identify. Obviously, without these hard positive samples, the model cannot be robust with occlusions and camera view changes. To mitigate the lack of hard samples, our framework further incorporates group pseudo labels, which include more hard samples, into the training process. The overall framework is illustrated in Figure \ref{fig:framework}.

\subsubsection{ Neighbor-based Groups Construction }
 A straight-forward approach to improve recall is to use a lower $\mu$, but it introduces too many random false-positive samples, which may bring the noise to model optimization. Therefore we construct neighbor-based groups, which have an inherent structure, to include more hard positive samples. At first, we make a basic assumption that the two samples have the same identity when they share a common neighbor. 
 At each training iteration, we use the GPP network to collect neighbors $\Omega_i$ of the given target sample $x_i^t$ and store the neighbors $\Omega_i$ into a neighbor memory $M_{neighbor}$.  After a training epoch, we merge two different samples $x^t_i, x^t_j$ into one group $G_k$ if $\Omega_i$ and $\Omega_j$ share a common neighbor according to our above assumption. However, the resulting group may contain too many samples. We control the group size by a maximum size $s$. To choose the maximum size $s$, we use DBSCAN \cite{DBLP:conf/kdd/EsterKSX96} on the memory bank $V$ to get the maximum cluster size as $s$. Finally, we merge all the samples into $z$ different groups $G=\{G_1,G_2,...,G_z\}$. By giving a pseudo label to each group, each target sample $x_i^t$ now has a group pseudo label $\hat{y}_i$. The whole group pseudo labels are denoted as $\hat{Y} = \{\hat{y}_i \vert 1 \leq i \leq N_t\}$.

 As a target sample, its group will contain more hard positive samples than its reliable neighbors due to the merging operation. However, the merged group is noisy because the predicted neighbors are not perfect and may contain some negative samples. Merging these negative samples' neighbors into a group will cause a group to contain multiple subgroups that correspond to multiple identities. It is the inherent structure of these \emph{noisy groups}.

 The above basic assumption is also applied in \cite{DBLP:journals/tmm/YeLYWLXCH16}. The two main differences are: (1) \cite{DBLP:journals/tmm/YeLYWLXCH16} needs two baseline methods to generate two different ranking lists, which is resource consuming. (2) Our method utilizes the assumption to construct group pseudo labels, while \cite{DBLP:journals/tmm/YeLYWLXCH16} utilizes the assumption to compute the similarity score and refine the final ranking list.

\begin{figure}[t]
	\centering
	\includegraphics[height=0.33\textwidth,width=0.45\textwidth]{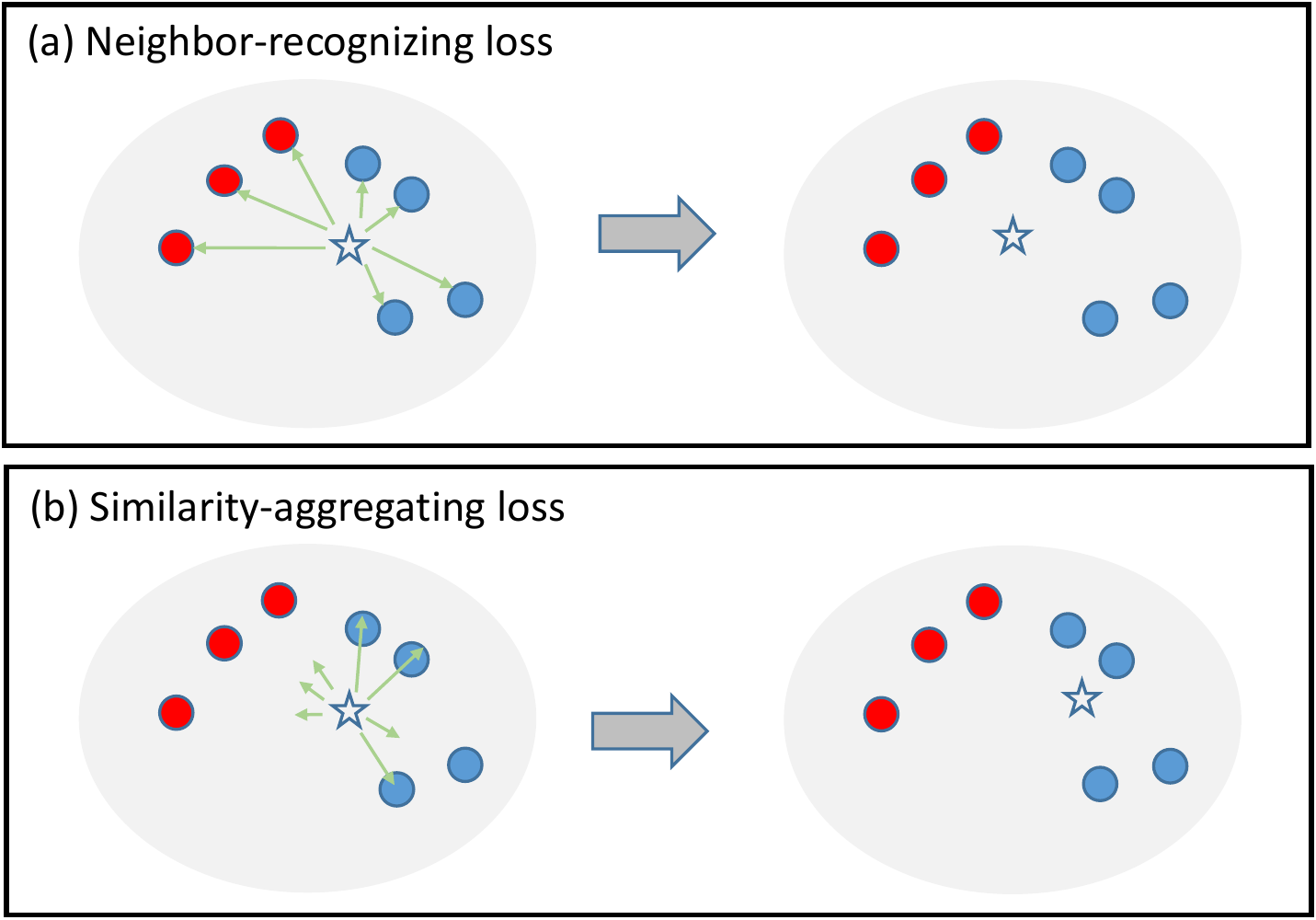}
	\caption{ Examples of the neighbor-recognizing loss and the similarity-aggregating loss on a noisy group that contains multiple subgroups. Different colors indicate different identities. (a): The neighbor-recognizing loss forces the embedding of the input sample to be the center of the whole group embeddings. (b): The similarity-aggregating loss pulls the input sample towards the most similar subgroup. ``$\star$'': the input sample. ``$\bullet$'': group members. }
	\label{fig:anaylze}
\end{figure}

\subsubsection{Optimize with Neighbor-based Groups}
A question may arise: \emph{can we directly utilize noisy groups $G$ by $L_{neighbor}$}? We first analyze the effect of this loss from the gradient view. Given the embedding $f_i$ of $x_i^t$, the gradient of the objective $L_{neighbor}$ respect to $f_i$ is
\begin{align}\label{eq:grad_neighbor}
  \frac{\partial L_{neighbor}}{\partial {f}_i} &= \frac{C}{\tau}(\sum_{j \in \Omega_i} (p_{ij} - \frac{w_{ij}}{C})v_j + \sum_{k\notin \Omega_i} p_{ik} v_k), \notag \\
  where\ C &= \sum_{j \in \Omega_i} w_{ij}
\end{align}
From the gradient, we observe that a large $p_{ij}$ will push $f_i$ far away from $v_j$, and a small $p_{ij}$ will pull $f_i$ close with $v_j$. When $p_{ik}=0$ and $p_{ij} = \frac{w_{ij}}{C}$, the gradient becomes $0$, which means that the optimal $f_i$ is at the center of the embeddings of $\Omega_i$. For reliable neighbors $\Omega_i$, using $L_{neighbor}$ can force $f_i$ equally similar to other neighbors' embeddings, which is desired because these neighbors are very likely to share the same identity with $x_i^t$. However, for a noisy group $G_i$, it may contain multiple subgroups that belong to different identities. If directly using $L_{neighbor}$ with $G_i$, the gradient will be severely affected by these subgroups.

To mitigate the influence of noisy subgroups, we hope that the loss function can enforce $f_i$ to be closer to the most similar subgroup than other subgroups. To achieve this goal, we propose to use a \textbf{similarity-aggregating loss} to utilize the group pseudo labels, which formulated as,
\begin{equation}\label{eq:loss_aggre}
   L_{aggre} =  - \log \sum\limits_{j} p_{ij}, j \in G_i \wedge j \neq i,
\end{equation}
where $G_i$ denotes the noisy group that the target image $x_i^t$ belongs to. Note that the summation operation is in the $\log()$, which is different from $L_{neighbor}$. This loss is also used in some studies \cite{DBLP:conf/eccv/WuEY18,DBLP:conf/iccv/ZhuangZY19}, but they only give an intuitive analysis. Here, we show the effect of $L_{aggre}$ from the gradient view. The gradient of the objective $L_{aggre}$ respect to $f_i$ is,
\begin{equation}\label{eq:grad_aggre}
   \frac{\partial L_{aggre}}{\partial {f}_i} = \frac{1}{\tau}(\sum_{j \in G_i} (p_{ij} - \frac{p_{ij}}{\sum_{k\in G_i} p_{ik}})v_j + \sum_{k\notin G_i} p_{ik} v_k ).
\end{equation}
It can be assured that $\sum_{k\in G_i} p_{ik} \leq 1$. Therefore the former item of the gradient is always negative. From the gradient, we can see that a larger $p_{ij}$ will enforce the $f_i$ to move closer to the $v_j$. Using $L_{aggre}$ with $G_i$, the $f_i$ will be pulled closer with the most similar subgroup than other subgroups. According to the assumption that similar embeddings are most likely to share the same identity if the embedding function is good enough, the most similar subgroup is most likely to share the same identity with $x_i^t$. Thus after training, the embeddings with the same identity can aggregate together, and we mitigate the influence of other subgroups. We show the different effects of these two loss functions in Figure. \ref{fig:anaylze}.

\begin{algorithm}[!ht]
  \caption{The joint learning framework}
  \label{alg:framework}
  \textbf{Require:} Source dataset $\mathcal{S}$, target dataset $\mathcal{T}$, training epochs $e_1$, $e_2$, total epochs $e_{all}$, embedding network $f(\theta)$, GPP network $g(\theta_1)$ \\
  \textbf{Ensure:} The best embedding network  $f(\theta)^*$. \\
  \begin{algorithmic}[1]
    \STATE Initialize: Neighbor Memory $M_{neighbor} = \{\{i\} \vert 1 \leq i \leq N_t\}$, group pseudo labels $\hat{Y} = \{i \vert 1 \leq i \leq N_t\}$
    \FOR{$i$ = 1 to $e_{all}$ }
      \STATE Use $M_{neighbor}$ to construct groups and assign $\hat{Y}$
      \FOR{mini-bacth $B_s \sim \mathcal{S}$ and $B_t \sim \mathcal{T}$}
        \STATE Calculate $L^{src}$ and $L_{gpp}$ using $B_s$
        \STATE $Loss = L^{src} + L_{gpp}$
        \IF{$i > e_1$ }
          \STATE Deploy $g(\theta_1)$ to predict the neighbors $\Omega_i$ of each sample in $B_t$;
          \STATE Use $\Omega_i$ to update $M_{neighbor}$;
          \STATE $Loss \mathrel{+}= L_{neighbor}$
        \ENDIF
        \IF{$i > e_2$ }
          \STATE $Loss \mathrel{+}= L_{aggre}$
        \ENDIF
        \STATE Optimize $f_{\theta}$ and $g(\theta_1)$ by $Loss$
      \ENDFOR
      \STATE Evaluate on the validation set $\rightarrow$ performance $P$;
      \IF{$P > P^*$}
        \STATE $P^* = P$, $f(\theta)^* = f(\theta)$
      \ENDIF
    \ENDFOR
  \end{algorithmic}
\end{algorithm}

\subsubsection{The Overall Loss Function}
It is a common practice to adopt both the classification loss and the triplet loss in fully supervised person re-ID \cite{DBLP:conf/cvpr/0004GLL019}. Viewing $L_{aggre}$ as the classification loss on the target domain, we also use the batch-hard triplet loss \cite{DBLP:journals/corr/HermansBL17} with the group pseudo label $\hat{Y}$, formulated as,
\begin{equation}
\begin{aligned}
L_{triplet}^{tgt} = [m + &\max_{p}||f_{i,a}^t-f_{i,p}^t||_2 \\
- & \min_{n}||f_{i,a}^t-f_{j,n}^t||_{2}]_+,
\end{aligned}
\end{equation}
where $f_{i,j}^t$ represents the embedding of the $j$-th image of the $i$-th pseudo identity in the batch, and $m=0.3$ denotes the distance margin.

Despite the performance drop, the model trained on the labeled source dataset still performs better than the ImageNet pre-trained model when directly tested on an unseen target dataset. Therefore, we follow \cite{DBLP:conf/cvpr/0004GLL019} to learn the basic discriminability of the target dataset using the labeled source dataset. Given a source sample $x_i^s$ and its identity $y_i^s$, we use the output of the embedding network to calculate softmax cross-entropy loss $L_{softmax}^{src}$ and hard-batch triplet loss $L_{triplet}^{src}$. $L_{triplet}^{src}$ is similar with $L_{triplet}^{tgt}$ but using $x_i^s$ and $y_i^s$. $L_{softmax}^{src}$ is in the form of,
\begin{equation}\label{eq:loss_src_softmax}
  L_{softmax}^{src} =  -\log p(y_i^s|x_i^s).
\end{equation}

Combining loss functions in the source branch and the target branch, the overall loss function used in jointly training strategy is formulated as,
\begin{equation}\label{eq:loss_all}
\begin{aligned}
L = & L_{softmax}^{src} + L_{triplet}^{src} + L_{gpp} + L_{neighbor} \\
& + L_{aggre} + L_{triplet}^{tgt},
\end{aligned}
\end{equation}
Note that we do not use the gradient of $L_{gpp}$ to optimize the embedding network $f(\theta)$. The whole optimize process is described in Algorithm \ref{alg:framework}. We train the model only with $L^{src}$ and $L_{gpp}$ at the first $e_1$ epochs. It can help the embedding network to learn basic discriminability and help the GPP network to classify neighbors. Then we add $L_{neighbor}$ after the first $e_1$ epochs to learn the discriminative information of the target domain. Finally, we add $L_{aggre}$ after the first $e_2$ epochs because the assumption of similarity-aggregation loss needs a good embedding function.

\begin{table*}[t]\setlength{\tabcolsep}{12pt}
  \caption{ Comparison with state-of-the-art unsupervised/semi-supervised methods on Market-1501 $\rightarrow$ DukeMTMC-ReID and Market-1501 $\rightarrow$ DukeMTMC-reID tasks. ``*'': semi-supervised method. }
  \label{tab:comp_sota}
  \centering
  \begin{tabular}{lrrrrrrrr}  
  \toprule
  \multirow{2}{*}{Methods}   & \multicolumn{4}{c}{DukeMTMC-ReID $\rightarrow$ Market-1501 } & \multicolumn{4}{c}{Market-1501 $\rightarrow$ DukeMTMC-ReID} \\
  \cmidrule{2-9}
  & mAP & R1   & R5   & R10  & mAP & R1   & R5   & R10  \\
  \midrule
  SPGAN+LMP \cite{DBLP:conf/cvpr/Deng0YK0J18} (CVPR18) & 26.7 & 57.7 & 75.8 & 82.4 & 26.2 & 46.4 & 62.3 & 68.0 \\
  UDAP \cite{DBLP:journals/pr/SongWZDZHW20} (arxiv18) & 53.7 & 75.8 & 89.5 & 93.2 & 49.0 & 68.4 & 80.1 & 83.5 \\
  ATNet \cite{DBLP:conf/cvpr/LiuZCH019} (CVPR19) & 25.6 & 55.7 & 73.2 & 79.4 & 24.9 & 45.1 & 59.5 & 64.2 \\
  ECN \cite{DBLP:conf/cvpr/Zhong0LL019} (CVPR19) & 43.0 & 75.1 & 87.6 & 91.6 & 40.4 & 63.3 & 75.8 & 80.4 \\
  UCDA-CCE \cite{DBLP:conf/iccv/QiWHZSG19} (ICCV19) & 34.5 & 64.3 & - & - & 36.7 & 55.4 & - & -\\
  PDA-Net \cite{DBLP:conf/iccv/LiLLW19} (ICCV19) & 47.6 & 75.2 & 86.3 & 90.2 & 45.1 & 63.2 & 77.0 & 82.5 \\
  PCB-PAST \cite{DBLP:conf/iccv/ZhangCSY19} (ICCV19) & 54.6 & 78.4 & - & - & 54.3 & 72.4 & - & - \\
  ECN+GPP \cite{DBLP:journals/corr/abs-1908-00485} (arxiv19)  & 63.8 & 84.1 & 92.8 & 95.4 & 54.4 & 74.0 & 83.7 & 87.4 \\
  SSG \cite{DBLP:conf/iccv/FuWWZSUH19} (ICCV19) & 58.3 & 80.0   & 90.0   & 92.4 & 53.4 & 73.0 & 80.6 & 83.2 \\
  ACT \cite{DBLP:conf/aaai/YangLZLSCGHJL20} (AAAI20) & 60.6 & 80.5 & - & - & 54.5 & 72.4 & - & - \\
  SSG$^{++}$* \cite{DBLP:conf/iccv/FuWWZSUH19} (ICCV19) & 68.7 & 86.2 & 94.6 & 96.5 & 60.3 & 76.0 & 85.8 & 89.3 \\
  MMT \cite{DBLP:conf/iclr/GeCL20} (ICLR20) & 71.2 & 87.7 & 94.9 & 96.9 & 65.1 & 78.0 & 88.8 & 92.5 \\
  \midrule
  Ours (ResNet-50) & \textbf{78.6} & \textbf{90.6} & \textbf{96.3} & \textbf{97.8}	& \textbf{67.9} & \textbf{81.3}	& \textbf{89.2} & \textbf{91.5}\\
  \midrule
  Ours (AGW) & 79.2 & 90.6 & 96.7 & 97.6	& 70.0 & 82.4	& 90.2 & 92.4 \\
  Ours (IBN-Net-a) & 81.1 & 91.4 & 97.1 & 98.0	& 68.3 & 81.2	& 89.3 & 91.7 \\
  \bottomrule
  \end{tabular}
\end{table*}
  
\begin{table}
  \caption{ Comparison with state-of-the-art unsupervised/semi-supervised methods on MSMT. ``*'': semi-supervised method. }
  \label{tab:results_msmt}
  \centering
  \begin{tabular}{lrrrrr}  
  \toprule
  \multirow{2}{*}{Methods} & \multirow{2}{*}{Source}  & \multicolumn{4}{c}{ MSMT } \\
  \cmidrule{3-6}
  && mAP & R1 & R5 & R10 \\
  \midrule
  Source only & Market & 5.2 & 15.3 & 23.6 & 28.1 \\
  PTGAN & Market & 2.9 & 10.2 & - & 24.4 \\
  ECN+GPP & Market & 15.2 & 40.4 & 53.1 & 58.7 \\
  SSG & Market & 13.2 & 31.6 & - & 49.6 \\
  SSG$^{++}$* & Market & 16.6 & 37.6 & - & 57.2 \\
  MMT & Market & \textbf{22.9} & \textbf{49.2} & \textbf{63.1} & \textbf{68.8} \\
  Ours & Market & 21.2 & 47.3 & 60.3 & 65.5 \\
  \midrule
  Source only & Duke & 7.2 & 21.8 & 32.1 & 37.7 \\
  PTGAN & Duke & 3.3 & 11.8 & - & 27.4 \\
  ECN+GPP & Duke & 16.0 & 42.5 & 55.9 & 61.5 \\
  SSG & Duke & 13.3 & 32.2 & - & 51.2 \\
  SSG$^{++}$* & Duke & 18.3 & 41.6 & - & 62.2 \\
  MMT & Duke & 23.3 & 50.1 & 63.9 & 69.8 \\
  Ours & Duke & \textbf{24.6} & \textbf{53.5} & \textbf{65.2} & \textbf{70.2} \\
  \bottomrule
  \end{tabular}
\end{table}

\begin{table*}[t]\setlength{\tabcolsep}{12pt}
  \caption{ Ablation studies of our proposed joint learning framework on Market-1501 and DukeMTMC-reID with ResNet-50 and IBN-Net-a backbones. \textbf{Supervised on target}: Model trained with labeled target data. \textbf{Source only}: Model trained with only labeled source data. \textbf{N}: Neighbor-recognizing loss for neighbor pseudo labels. \textbf{S}: Similarity-aggregating loss for group pseudo labels. \textbf{T}: Target triplet loss for group pseudo labels. }
  \label{tab:ablation}
  \centering
  \begin{tabular}{lrrrrrrrr}  
  \toprule
  \multirow{2}{*}{Methods (ResNet-50)}   & \multicolumn{4}{c}{DukeMTMC-ReID $\rightarrow$ Market-1501 } & \multicolumn{4}{c}{Market-1501 $\rightarrow$ DukeMTMC-ReID} \\
  \cmidrule{2-9}
  & mAP & R1   & R5   & R10  & mAP & R1   & R5   & R10  \\
  \midrule
  Supervised on target & 83.9 & 94.6 & 98.1 & 99.0 & 72.2 & 85.2 & 92.9 & 94.8 \\
  Source only & 28.2 & 58.0 & 73.1 & 78.6 & 25.0 & 41.3 & 57.5 & 64.1 \\
  \midrule
  ECN+GPP \cite{DBLP:journals/corr/abs-1908-00485} & 63.8 & 84.1 & 92.8 & 95.4 & 54.4 & 74.0 & 83.7 & 87.4 \\
  Ours w/ N (baseline) & 64.7 & 81.6 & 92.6 & 95.3 & 62.6 & 77.1 & 86.4 & 89.6 \\
  \midrule
  Ours w/ N+S & 74.9 & 88.5 & 94.8 & 96.7 & 67.6 & 79.8 & 89.1 & 91.7 \\
  Ours w/ N+T & 70.0 & 84.9 & 93.5 & 95.8 & 64.2 & 78.1 & 87.7 & 90.1 \\
  Ours w/ N+S+T & \textbf{78.6} & \textbf{90.6} & \textbf{96.3} & \textbf{97.8}	& \textbf{67.9} & \textbf{81.3}	& \textbf{89.2} & \textbf{91.5}\\
  Ours w/ S+T & 59.1 & 82.2 & 90.8 & 93.5 & 59.2 & 74.0 & 83.5 & 87.2 \\
  \bottomrule
  \end{tabular}
\vspace{1pt}  
  \begin{tabular}{lrrrrrrrr}  
    \toprule
    \multirow{2}{*}{Methods (IBN-Net-a)}   & \multicolumn{4}{c}{DukeMTMC-ReID $\rightarrow$ Market-1501 } & \multicolumn{4}{c}{Market-1501 $\rightarrow$ DukeMTMC-ReID} \\
    \cmidrule{2-9}
    & mAP & R1   & R5   & R10  & mAP & R1   & R5   & R10  \\
    \midrule
    Supervised on target & 84.8 & 94.3 & 98.0 & 98.7 & 74.4 & 86.8 & 93.4 & 95.2 \\
    Source only & 31.0 & 61.1 & 75.9 & 80.8 & 32.5 & 51.0 & 65.3 & 71.2 \\
    \midrule
    Ours w/ N & 70.9 & 86.0 & 94.6 & 96.5 & 64.9 & 78.1 & 88.2 & 90.4 \\
    \midrule
    Ours w/ N+S & 80.5 & 91.2 & 96.5 & 97.9 & 67.8 & 80.6 & 89.2 & 90.9 \\
    Ours w/ N+T & 75.0 & 88.3 & 94.9 & 96.7 & 65.3 & 78.8 & 87.9 & 90.0 \\
    Ours w/ N+S+T & \textbf{81.1} & \textbf{91.4} & \textbf{97.1} & \textbf{98.0}	& \textbf{68.3} & \textbf{81.2}	& \textbf{89.3} & \textbf{91.7} \\
    Ours w/ S+T & 66.2 & 86.2 & 93.2 & 95.3 & 61.6 & 76.2 & 85.7 & 88.9 \\
    \bottomrule
    \end{tabular}
\end{table*}

\section{Experiments}
\subsection{Datasets and Evaluation Protocol}
We evaluate our method on three large-scale person re-identification (re-ID) datasets: Market-1501 \cite{DBLP:conf/iccv/ZhengSTWWT15}, DukeMTMC-reID \cite{DBLP:conf/eccv/RistaniSZCT16,DBLP:conf/iccv/ZhengZY17} and MSMT \cite{DBLP:conf/cvpr/WeiZ0018}. 

\textbf{Market1501} includes 32,668 images of 1,501 different identities captured by 6 cameras on the university campus. There are 12,936 images of 751 identities for training and 19,732 images of 750 identities for testing.

\textbf{DukeMTMC-reID} contains 36,411 images captured by 8 cameras in winter. The training set has 16,522 images of 702 different identities. The testing set contains 2,228 query images and 17,661 gallery images of another 702 identities.

\textbf{MSMT} is the largest person re-ID dataset. It contains 126,411 images captured by 15 cameras. Among them, 32,621 images of 1041 identities are used as the training set. The testing set contains 93,820 images of 3,060 identities.

\textbf{Evaluation Protocol.} We use the cumulative matching characteristic (CMC) and the mean Average Precision (mAP) as our evaluation metrics. For a fair comparison, we do not use any re-ranking techniques\cite{DBLP:conf/cvpr/ZhongZCL17} and evaluate under the single query mode.

\subsection{Implementation Details}
We adopt ResNet-50 \cite{DBLP:conf/cvpr/HeZRS16} and IBN-Net-a \cite{DBLP:conf/eccv/PanLST18} as our backbone network, which are both pre-trained on ImageNet \cite{DBLP:conf/cvpr/DengDSLL009}. We set the stride of the last residual block to $1$ and add a batch normalization layer \cite{DBLP:conf/icml/IoffeS15} after the global average pooling (GAP) layer as \cite{DBLP:conf/cvpr/0004GLL019}. We resize the input images to $256 \times 128$. During training, we use random flipping, random cropping, and random erasing \cite{DBLP:conf/aaai/Zhong0KL020} for data augmentation. For the need of calculating batch-hard triplet loss, we randomly select $8$ identities and randomly sample $4$ images for each identity to form the mini-batch for source images. The sampling is similar for target images, except we use the noisy label $\hat{y_i}$ as the pseudo label. We use Adam as optimizer to optimize our model for $70$ epochs. The initial learning rate is $1.25 \times 10^{-4}$ and divided by $10$ after the first $40$ epochs. 

We empirically set $e_1 = 5$ and $e_2 = 10$. We select $k=200$ nearest-neighbors as candidate neighbors for the GPP network. In the testing stage, we use the L2-normalized output of the embedding network as the image embedding. The settings about the GPP network are kept the same as \cite{DBLP:journals/corr/abs-1908-00485}.

\subsection{Comparison with State-of-the-art Methods}
We compare our method with state-of-the-art UDA methods on Market-1501, DukeMTMC, and MSMT17. Our method can achieve state-of-the-art performance. These results demonstrate the effectiveness of our joint learning framework.

We report the results on Market-1501 and DukeMTMC-reID in Table. \ref{tab:comp_sota}. The source dataset is Market-1501 when tested on DukeMTMC, and vice versa. We compare with two reducing domain gap methods: SPGAN+LMP \cite{DBLP:conf/cvpr/Deng0YK0J18} and ATNet \cite{DBLP:conf/cvpr/LiuZCH019}, three clustering-based pseudo label methods: UDAP \cite{DBLP:journals/pr/SongWZDZHW20}, SSG \cite{DBLP:conf/iccv/FuWWZSUH19} and PCB-PAST \cite{DBLP:conf/iccv/ZhangCSY19}, two neighbor-based pseudo label methods: ECN \cite{DBLP:conf/cvpr/Zhong0LL019} and ECN+GPP \cite{DBLP:journals/corr/abs-1908-00485}, a semi-supervised method SSG$^{++}$ \cite{DBLP:conf/iccv/FuWWZSUH19}, two label noise about method ACT \cite{DBLP:conf/aaai/YangLZLSCGHJL20} and MMT \cite{DBLP:conf/iclr/GeCL20}. As shown in Table. \ref{tab:comp_sota}, these reducing domain gap methods fail to obtain competitive results because they ignore the discriminative information on the target domain. Compared with other pseudo label methods, our method significantly outperforms them. In particular, by adopting the ResNet-50 backbone, we compare with ECN+GPP \cite{DBLP:journals/corr/abs-1908-00485}. Our results achieve $14.8\%$ and $13.5\%$ mAP improvement on Market1501 and DukeMTMC. Compared with MMT, our method outperforms with $7.4\%$ and $2.8\%$ mAP. Moreover, compared with the semi-supervised domain adaptation method SSG$^{++}$, our method still achieves $9.9\%$ and $7.6\%$ mAP improvement when tested on the Market-1501 and DukeMTMC dataset. Besides, with the help of IBN-Net-a \cite{DBLP:conf/eccv/PanLST18} and AGW \cite{DBLP:journals/corr/abs-2001-04193}, our method can be very close to fully-supervised learning performances shown in \ref{tab:ablation}. 

In Table. \ref{tab:results_msmt}, we further report our results on MSMT17 that is the most challenging dataset. A similar improvement can be observed. Using ResNet-50 as backbone, we achieve mAP= $24.6\%$ and rank-1 accuracy= $53.5\%$ when trained on DukeMTMC-reID, and mAP= $21.2\%$ and rank-1 accuracy= $47.3\%$ when trained on Market-1501. Our results surpass the semi-supervised method SSG$^{++}$, which further demonstrates the effectiveness of our proposed method.

\subsection{Ablation Studies}
\subsubsection{Performance of Supervised Learning}
In Table. \ref{tab:ablation}, we report the performance of our supervised counterpart and the direct transfer method. Both of them use only $L^{src}$ in our framework. We can find that the supervised method can achieve high performance if the training domain and the testing domain are the same. However, the performance severely drops when tested on an unseen domain. For example, taking Market-1501 as the source domain and DukeMTMC-reID as the target domain, the model trained on the source domain can achieve $83.9\%$ mAP when tested on the source domain. However, it drops to $25.0\%$ mAP when tested on the target domain. A similar performance drop can be observed if exchanging the source domain and the target domain.

\begin{table*}\setlength{\tabcolsep}{12pt}
  \caption{Ablation studies of loss functions with different pseudo labels on ResNet-50. ``NR'': Neighbor-recognizing loss. ``SA'': Similarity-aggregating loss. ``CE'': Cross-entropy loss. }
  \label{tab:exchange}
  \centering
  \begin{tabular}{lrrrrrrrrr}  
  \toprule
  \multirow{2}{*}{Loss} & \multirow{2}{*}{Pseudo label}  & \multicolumn{4}{c}{DukeMTMC-ReID $\rightarrow$ Market-1501 } & \multicolumn{4}{c}{Market-1501 $\rightarrow$ DukeMTMC-ReID} \\
  \cmidrule{3-10}
  && mAP & R1   & R5   & R10  & mAP & R1   & R5   & R10  \\
  \midrule
  NR & group & 72.9	& 86.3 & 95.0 & 96.8 & 54.0 & 70.6 & 84.0 & 88.1 \\
  CE & group & 59.7	& 79.5 & 91.4 & 94.3 & 50.8 & 68.5 & 81.4 & 85.3 \\
  SA & group & \textbf{78.6} & \textbf{90.6} & \textbf{96.3} & \textbf{97.8}	& \textbf{67.9} & \textbf{81.3}	& \textbf{89.2} & \textbf{91.5} \\
  \midrule
  NR & neighbor & \textbf{78.6} & \textbf{90.6} & \textbf{96.3} & \textbf{97.8}	& \textbf{67.9} & \textbf{81.3}	& \textbf{89.2} & \textbf{91.5} \\
  SA & neighbor & 50.9 & 74.9 & 85.5 & 89.6 & 49.1 & 64.0 & 76.2 & 81.7 \\
  \bottomrule
  \end{tabular}
\end{table*}

\subsubsection{Effectiveness of Neighbor-based Groups}
We perform several ablation studies to prove the necessity of group pseudo labels. Because our neighbor-recognizing loss follows ECN+GPP \cite{DBLP:journals/corr/abs-1908-00485}, it can be viewed as a reproduction of this paper. As shown in Table. \ref{tab:ablation}, our reproduction (``Ours w/ N'') outperforms the performance of ECN+GPP $0.9\%$ and $8.2\%$ mAP when tested on Market-1501 and DukeMTMC-reID. For a fair comparison, we use it as our \textbf{baseline}. 

First, we show the effectiveness of the similarity-aggregating loss on noisy groups by integrating it with our \textbf{baseline}, which is denoted as ``Ours w/ N+S''. In Table. \ref{tab:ablation}, we observe significant improvement by adding the similarity-aggregating loss. Specifically, using ResNet-50 as the backbone, the mAP is improved by $10.2\%$ and $5.0\%$ when tested on Market-1501 and DukeMTMC-reID, respectively. A similar improvement can be observed using IBN-Net-a as the backbone. These improvements demonstrate that neighbor-based groups can discover more discriminative information than reliable neighbors.

Next, we validate the effectiveness of the triplet loss on noisy groups. As reported in Table. \ref{tab:ablation}, ``Ours w/N+T'' achieve a rank-1 accuracy of $84.9\%$ when tested on Market-1501, which is $3.3\%$ higher than the \textbf{baseline}. This improvement again demonstrates the effectiveness of neighbor-based groups.

Furthermore, our method would obtain more improvement when incorporating the similarity-aggregating loss and the target triplet loss into the \textbf{baseline}. ``Ours w/N+S+T'' can always achieve the highest mAP on Market-1501 and DukeMTMC-reID when using ResNet-50 or IBN-Net-a as the backbone. This achievement indicates that the model learns more discriminative information from more hard positive samples.

\begin{table}\setlength{\tabcolsep}{12pt}
  \caption{Ablation studies of the number of unlabeled samples for adaptation. The percentage denotes the number of target domain data. }
  \label{tab:percent}
  \centering
  \begin{tabular}{rrrrr}  
  \toprule
  \multirow{2}{*}{Percent}  & \multicolumn{2}{c}{D $\rightarrow$ M } & \multicolumn{2}{c}{M $\rightarrow$ D} \\
  \cmidrule{2-5}
  & mAP & R1   & mAP & R1 \\
  \midrule
  100\% & \textbf{78.6}	& \textbf{90.6} & \textbf{67.9} & \textbf{81.3} \\
  75\% & 73.4	& 87.7 & 66.4 & 79.2 \\
  50\% & 65.3	& 82.7 & 62.9 & 77.0 \\
  25\% & 50.0	& 71.0 & 53.5 & 69.5 \\
  0\% & 28.2 & 58.0 & 25.0 & 41.3 \\
  \bottomrule
  \end{tabular}
\end{table}

\begin{table}\setlength{\tabcolsep}{12pt}
  \caption{Ablation studies of $e_1$ and $e_2$. }
  \label{tab:epoch}
  \centering
  \begin{tabular}{lrrrrr}  
  \toprule
  \multirow{2}{*}{$e_1$} & \multirow{2}{*}{$e_2$}  & \multicolumn{2}{c}{D $\rightarrow$ M } & \multicolumn{2}{c}{M $\rightarrow$ D} \\
  \cmidrule{3-6}
  && mAP & R1   & mAP & R1 \\
  \midrule
  0 & 5 & 77.4	& 90.1 & \textbf{68.4} & 80.9  \\
  5 & 10 & \textbf{78.6}	& \textbf{90.6} & 67.9 & \textbf{81.3}  \\
  10 & 20 & 76.9	& 89.5 & 67.6 & 80.7  \\
  20 & 30 & 75.6	& 89.0 & 66.2 & 79.7  \\
  30 & 40 & 72.5	& 89.2 & 63.5 & 77.5  \\
  \bottomrule
  \end{tabular}
\end{table}

\subsubsection{Effectiveness of Similarity-Aggregating Loss}
To experimentally show the effectiveness of similarity-aggregating loss, we conduct ablation studies by exchanging the \textbf{similarity-aggregating loss} and the \textbf{neighbor-recognizing loss}. In Table. \ref{tab:exchange}, we first verify similarity-aggregating loss on our best model by using the neighbor-recognizing loss for group pseudo labels. Its mAP is $5.7\%$ lower on Market1501 and $13.9\%$ lower on DukeMTMC-reID than using similarity-aggregating loss. Next, we demonstrate the effectiveness of neighbor-recognizing loss. As shown in Table. \ref{tab:exchange}, using similarity-aggregating loss for reliable neighbors causes severe performance drop. The reason for the drop is that similarity-aggregating loss aggregates similar samples in reliable neighbors and pushes other samples far away. Pushing too many hard positive samples away will undoubtedly influence the optimization of the model.

Another choice to utilize group pseudo labels is to use cross-entropy loss as most cluster-based UDA methods used \cite{DBLP:conf/iccv/ZhangCSY19,DBLP:conf/iccv/FuWWZSUH19}. The form of this loss is similar to $L_{softmax}^{src}$ but uses $\hat{y}_i$ as the label. From Table. \ref{tab:exchange}, we observe significant $18.9\%$ and $17.1\%$ mAP drops on Duke-to-Market and Market-to-Duke tasks, respectively. The reason is that the effect of the cross-entropy loss is similar to $L_{neighbor}$, and it cannot tackle the noise of the group pseudo labels.

\subsubsection{Necessity of Neighbor Pseudo labels}
We show the effectiveness of neighbor-based groups in the above ablation studies, but can we use only neighbor-based groups? The neighbors are the foundation of the model optimization and cannot be ignored. We know the assumption of the similarity-aggregation loss is a good embedding function. Without neighbor pseudo labels, it is not easy to get such an embedding function. In Table. \ref{tab:ablation}, we only use group pseudo labels to optimize the model, denoted as ``Ours w/ S+T''. The performance is much lower than the ``Ours w/ N+S+T''.

\subsubsection{Analysis of more constraints}
Can we add more constraints to reduce the noise of the merged groups? In Table \ref{tab:constraints}, we try two different constraints that (a) the two samples have the same identity when they share \textbf{many} common neighbors and (b) vote the final neighbors by the neighbors in this epoch and previous epoch. However, neither can achieve better performance. We argue that this is a precision-recall trade-off problem. With more constraints added, the recall will also degrade. We may need some better constraints to reduce noise without affecting the recall.

\subsection{Algorithm Analysis}
\subsubsection{Analysis of The Number of Unlabeled Samples}
To show how the performance changes as the number of unlabeled target samples for adaptation is changed, we report the performances in Table. \ref{tab:percent} when different ratios of unlabeled data are used. When using $75\%$ data, we can still get better performance than some state-of-the-art methods. Even when using $25\%$ data, our method can always improve the model.

\subsubsection{Analysis of $e_1$ and $e_2$}
In Table \ref{tab:epoch}, we report the impact of $e_1$ and $e_2$. Assigning lower values to $e_1$ and $e_2$
starts discovering discriminative information of the target domain early, which produces better results. And the change of $e_1$ and $e_2$ has slight influence to the performance unless assigning too large values or too small values. The best results are obtained at $e_1 = 5$ and $e_2 = 10$.

\begin{table}\setlength{\tabcolsep}{12pt}
  \caption{ Ablation studies of different constraints. }
  \label{tab:constraints}
  \centering
  \begin{tabular}{lrrrrr}  
  \toprule
  \multirow{2}{*}{method} & \multicolumn{2}{c}{D $\rightarrow$ M } & \multicolumn{2}{c}{M $\rightarrow$ D} \\
  \cmidrule{2-5}
  & mAP & R1   & mAP & R1 \\
  \midrule
  Our & \textbf{78.6}	& \textbf{90.6} & \textbf{67.9} & \textbf{81.3}  \\
  Constraint (a) & 76.0	& 88.5 & 66.8 & 80.0  \\
  Constraint (b) & 77.1	& 89.6 & 67.3 & 79.2  \\
  \bottomrule
  \end{tabular}
\end{table}

\begin{table}\setlength{\tabcolsep}{12pt}
  \caption{ Precision, recall, and F1-score of different methods' pseudo labels on Duke-to-Market task. }
  \label{tab:pseudo}
  \centering
  \begin{tabular}{lrrr}  
  \toprule
  Method & precision & recall & F1 \\
  \midrule
  MMT\cite{DBLP:conf/iclr/GeCL20} & 60.5 & 94.5 & 73.8   \\
  SpCL\cite{DBLP:conf/nips/Ge0C0L20} & 75.6 & 82.4 & 78.9   \\
  Neighbor (our w/o SA) & 79.1 & 52.9 & 63.4 \\
  Group (our w/o SA) & 75.9  & 68.5 & 72.0  \\
  Neighbor (our) & 83.1 & 68.6 & 75.2 \\
  Group (our) & 81.2  &  80.6 & 80.9  \\
  \bottomrule
  \end{tabular}
\end{table}

\subsubsection{Analysis of pseudo labels} 
Compared with MMT\cite{DBLP:conf/iclr/GeCL20} and SpCL\cite{DBLP:conf/nips/Ge0C0L20}, we show the precision, recall, and F1 score of their pseudo labels in Table. \ref{tab:pseudo}. As shown in Table. \ref{tab:pseudo}, our neighbor pseudo labels' precision surpasses MMT and SpCL. Our group pseudo labels can achieve higher recall than neighbor pseudo labels while the precision does not drop dramatically. MMT's pseudo labels have high recall but suffer low precision. SpCL's pseudo labels have similar recall and precision with our group pseudo labels but ignore the high precision pseudo labels. Besides, using similarity-aggregating loss can improve both the recall and precision of neighbors and groups because these discovered hard positive samples can improve the quality of the embedding function.

\section{Conclusion}
In this paper, we propose a joint learning framework to mitigate the problem of missing hard samples of neighbor-based UDA methods for re-ID. Specifically, we use two complementary pseudo labels, \textit{i.e.} high precision neighbor pseudo labels and high recall group pseudo labels to optimize the embedding network. With the help of group pseudo labels, the model can learn more discriminative information of the target domain. Furthermore, we analyze the inherent structure of the merged groups and introduce a similarity-aggregating loss to mitigate the influence of the noisy samples in the group. Extensive experimental results demonstrate that the performance of our method outperforms state-of-the-art methods significantly and is close to the fully supervised counterpart.


%





\ifCLASSOPTIONcaptionsoff
  \newpage
\fi



\bibliographystyle{IEEEtran}
\bibliography{Bibliography}
\end{document}